\documentclass[runningheads]{llncs}

\usepackage[year=2026]{accv}

\usepackage{accvabbrv}

\usepackage{graphicx}
\usepackage{booktabs}
\usepackage{float}
\usepackage{algorithm}
\usepackage{algpseudocode}
\usepackage{amsmath}

\usepackage[accsupp]{axessibility}  

\usepackage[breaklinks,colorlinks,citecolor=accvblue]{hyperref}

\hypersetup{
  pdftitle={CARE: Camera-Residual Reserves for First Sightings in Adaptive LiDAR Sensing},
  pdfauthor={Jiachen Gong and Yun Li and Ehsan Javanmardi and Wencan Mao and Manabu Tsukada},
  pdfsubject={Adaptive LiDAR sensing; sensing budget allocation; 3D object detection},
  pdfkeywords={adaptive LiDAR sensing, sensing budget allocation, camera-guided scanning,
               3D object detection, autonomous driving},
  pdflang={en-US},
}

\usepackage{orcidlink}

\begin{document}

\title{CARE: Camera-Residual Reserves for First Sightings in Adaptive LiDAR Sensing}

\titlerunning{CARE: Camera-Residual Reserves}

\author{Jiachen Gong\inst{1}\orcidlink{0009-0007-7856-9925} \and
Yun Li\inst{1}\orcidlink{0009-0002-7824-7751} \and
Ehsan Javanmardi\inst{1}\orcidlink{0000-0003-0337-115X} \and
Wencan Mao\inst{2}\orcidlink{0000-0002-3971-8314} \and
Manabu Tsukada\inst{1}\orcidlink{0000-0001-8045-3939}}

\authorrunning{J.~Gong et al.}

\institute{The University of Tokyo, Tokyo, Japan\\
\email{\{jiachengong,li-yun,ejavanmardi,mtsukada\}@g.ecc.u-tokyo.ac.jp} \and
National Institute of Informatics, Tokyo, Japan\\
\email{wencan\_mao@nii.ac.jp}}


\maketitle

\begin{abstract}
  Adaptive LiDAR scanning, which concentrates a limited sensing budget on informative regions of interest based on past object tracks, is becoming a key enabler in autonomous driving by lowering data volume while maintaining detection accuracy.
  However, existing scanning policies face three key challenges.
  First, history-driven approaches depend heavily on past object tracks, leading to delayed or missed detection of unseen objects.
  Second, methods leveraging random or uniform sampling outside the predicted regions have no awareness of where new objects appear.
  Third, camera-guided alternatives spend their budget on all camera detections, resampling objects that are already covered, which costs overall recall in crowded scenes and detection range when budgets are scarce.
  To address these limitations, this paper introduces the CAmera-REsidual reserve (CARE), a training-free allocation rule that reserves part of a fixed ray budget for the directions of current camera detections that the track forecasts cannot explain, while the remaining budget follows the base history policy and unused reserve returns to a shared random floor.
  The paper makes three key contributions: First, a leakage-free ray-budget evaluation on nuScenes (150 scenes, 4{,}148 events) that measures the first-sighting loss of history-driven scanning, with a strict-causal variant using the preceding keyframe.
  Second, CARE raises first-sighting recall by 5.2, 5.2, and 4.3 points at 10\%, 20\%, and 35\% budgets over the history policy with paired intervals excluding zero; the camera cue drives this gain, and the first-sighting versus overall trade-off is reported openly as a budget-dependent Pareto choice.
  Third, a safety-bounded forgetting module that releases budget from receding or static tracks beyond a speed-dependent guard distance; at tight budgets forgetting without the guard significantly harms near-field recall, and the guard is what keeps forgetting safe at the tightest budget.
  The pipeline runs end to end on a real vehicle and, in a closed-loop simulation, detects an occluded pedestrian earlier and brakes more reliably than history-driven scanning.
\end{abstract}

\section{Introduction}
\label{sec:intro}

LiDAR is the main range sensor of most autonomous driving stacks~\cite{caesar2020nuscenes,yin2021centerpoint}, but a dense scan of the full field of view at every frame spends much of the sensing budget on empty road, building walls, and objects the system already tracks with high confidence. Temporal cues are now standard inside detection networks~\cite{wang2023streampetr,lin2022sparse4d,liu2023sparsebev}, and adaptive LiDAR scanning brings the same idea to the sensor itself: it concentrates a limited ray budget on regions of interest that are predicted from the recent past, and recent systems report large savings in scanning density~\cite{memslidar2020,shomer2023sampledepth}, the most recent at almost no loss in detection accuracy~\cite{adaptivelidar2026}.

However, existing history-driven scanning policies face three key challenges. First, an object that has no track yet produces no region of interest, so pedestrians and vehicles that newly enter the field of view receive the least sensing budget; we call this failure \emph{first-sighting under-allocation}. Second, the usual fallback outside the predicted regions is random or uniform sampling, which is not directed at where new objects appear and, as our experiments show, can even fall below the plain history policy. Third, camera cues are either fused inside the detector~\cite{bai2022transfusion,liu2023bevfusion,yan2023cmt} or steer the scan without asking whether the tracker already covers the object~\cite{memslidar2020}, so part of the budget goes to objects that need no extra sensing.

To address these limitations, this paper introduces the CAmera-REsidual reserve (CARE), a training-free allocation rule built on one observation: a camera detection is useful for scan allocation only to the extent that the temporal memory cannot already explain it. At each decision time CARE projects the current track forecasts into the camera image, removes every detection that a same-class forecast already explains, and reserves part of the ray budget for the viewing directions of the remaining detections. The rest of the budget follows the base history policy, and any budget the reserve leaves unused is filled by a shared random floor, so the mechanism falls back to the history policy when the camera sees nothing new or fails entirely. A safety-bounded forgetting module (SBF) further releases budget from tracked objects that are receding or static and outside a speed-dependent guard distance, and hands the released cells to the same reserve; the guard is what keeps this forgetting safe at the tightest budget.

The main contributions of this paper are summarized as follows.
\begin{enumerate}
\item A leakage-free evaluation protocol that measures first-sighting under-allocation on a fixed angular ray-cell budget: cells are chosen before the returns are observed, all policies share the same detector, budget, and random floor, and first sightings are scored with paired per-scene bootstrap intervals; a strict-causal variant delays the camera cue by one keyframe.
\item The CARE allocation rule. On nuScenes validation (150 scenes, 4{,}148 events), CARE raises first-sighting recall by 5.2, 5.2, and 4.3 points at 10\%, 20\%, and 35\% budgets over the history policy, with paired confidence intervals excluding zero. An all-camera control attributes this gain to the camera cue rather than the history state, and residual filtering directs the reserve only at objects the memory cannot explain, spending nothing on tracked ones. The first-sighting versus overall trade-off is reported as a budget-dependent Pareto choice.
\item The safety-bounded forgetting module, which combines time-decay forgetting of low-risk tracks with a speed-dependent guard distance and a localization-noise margin, all set from published traffic rules and models rather than tuning. On the ablation subset at the tightest budget, forgetting without the guard loses 1.8 points of near-field recall against CARE without forgetting, with an interval excluding zero, while the guarded default shows no significant loss. The pipeline is further demonstrated on a real vehicle and in a closed-loop simulation study.
\end{enumerate}
CARE is further compared against policy-level re-implementations of recent and standard alternatives, including history-driven scanning~\cite{adaptivelidar2026}, an uncertainty-driven reserve~\cite{ancha2020lightcurtains}, and beam reduction~\cite{wei2022lidardistill}, all under the same allocator, budget, and detector.

\begin{figure}[H]
\centering
\includegraphics[width=0.94\textwidth]{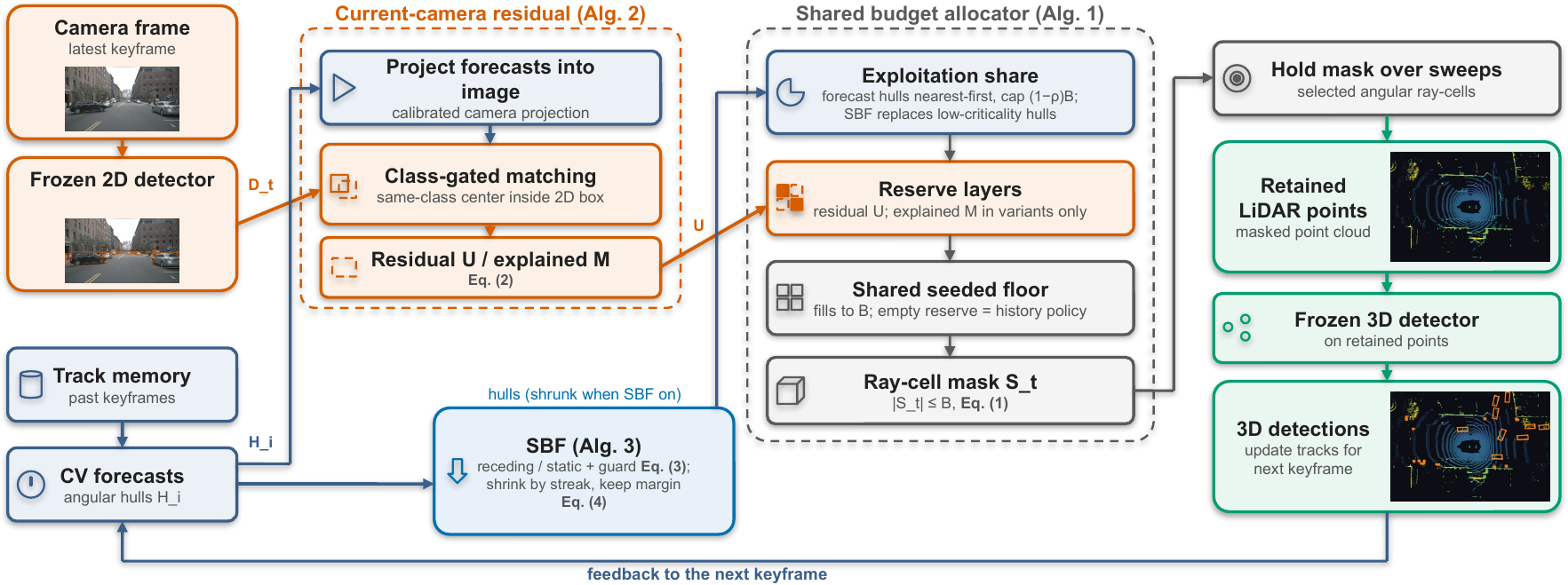}
\caption{Overview of CARE at one keyframe. Camera detections $D_t$ that the projected track forecasts (angular hulls $H_i$) cannot explain form the residual layer $U$ (Algorithm~\ref{alg:residual}), which fills the reserve of the shared allocator (Algorithm~\ref{alg:allocator}) ahead of the seeded floor; the SBF stage (Algorithm~\ref{alg:sbf}) shrinks the hulls of low-criticality tracks. The output mask $S_t$ is held over the sweeps, and the 3D detections update the track memory.}
\label{fig:arch}
\end{figure}

\section{Related Work}
\label{sec:related}

\subsubsection{Adaptive and temporal LiDAR scanning.}
Programmable LiDAR can spread its rays unevenly over the field of view, and early systems steered this freedom with image cues~\cite{memslidar2020}, a scene prior~\cite{shomer2023sampledepth}, or human visual attention~\cite{scari2025hybrid}. The most recent and strongest line derives the regions of interest from the tracks the perception system already holds: a history-aware predictor with a learned mask generator concentrates dense scanning on forecast object regions, cutting LiDAR energy by more than 65\% at comparable accuracy~\cite{adaptivelidar2026}, and forecast positions have been fed directly as detection queries~\cite{zhang2026pred}. These history-driven policies share one structural property: every dense region traces back to an object the system has already seen, so an object at its first appearance, which has no track yet, is given no priority. CARE keeps this history policy as its base and adds the one cue that the history state cannot contain, a current-frame camera detection, so that newly entering objects are covered.

\subsubsection{Active perception with uncertainty.}
A second family points the sensor where the current estimate is least certain: programmable light curtains place a thin sensing surface at the least certain depth, with safety-envelope variants tracking the closest possible obstacle~\cite{ancha2020lightcurtains,ancha2021envelopes}, building on uncertainty estimation in deep vision models~\cite{kendall2017uncertainties}. Uncertainty is a useful cue, but it can only be formed where the sensor has already returned points, so an object that has never been measured produces no uncertainty to react to. Our experiments include a reserve of this kind, and it reaches the highest overall recall of any policy we test, yet it trails CARE at every budget on first sightings, exactly the gap between revisiting measured but uncertain regions and directing sensing toward new objects.

\subsubsection{Post-acquisition sampling and beam reduction.}
A different line reduces LiDAR data after the full scan is captured. Farthest point sampling is the standard geometric choice~\cite{qi2017pointnetpp}, learned samplers keep the task-relevant points~\cite{lang2020samplenet,zhang2022iassd}, beam reduction drops whole scan lines and is the common protocol for cross-density studies~\cite{wei2022lidardistill}, and related methods prune tokens or voxels inside the network~\cite{rao2021dynamicvit,zhao2023ada3d}. All of these subsample a cloud that has already been acquired, so they save downstream computation but not a single ray of sensing; we keep random and farthest point sampling as such references, and adapt beam reduction and a learned scorer into causal scan policies for comparison.

\subsubsection{Camera-LiDAR fusion.}
Modern 3D detectors combine camera and LiDAR features inside the network~\cite{bai2022transfusion,liu2023bevfusion,chen2023futr3d,yan2023cmt,xie2023sparsefusion}, and the strongest systems on nuScenes are multi-modal~\cite{caesar2020nuscenes}. This fusion happens after the scan pattern is already fixed, so the camera improves what is done with the returned points but never changes where the LiDAR looks. CARE uses the camera one step earlier, as the cue that decides where the next rays go; it reads only boxes from a frozen 2D detector, leaves the downstream detector untouched, and could sit in front of a fusion model as well.

\section{Problem Formulation}
\label{sec:problem}

\subsubsection{Budgeted ray-cell scanning.}
We discretize the LiDAR field of view into an angular grid $\mathcal{G}$ of $A$ azimuth columns and $E$ elevation rows, giving $N = A \cdot E$ firing cells. A scanning policy selects a subset $S_t \subseteq \mathcal{G}$ at decision time $t$ under a budget
\begin{equation}
|S_t| \le B, \qquad B = \mathrm{round}(\beta N),
\label{eq:budget}
\end{equation}
where $\beta \in (0,1]$ is the budget fraction. Only points whose viewing direction falls in a selected cell are retained, and the masked cloud is passed to a frozen 3D detector. Selection is \emph{leakage-free}: $S_t$ is fixed before the returns of frame $t$ are observed, so a policy may use the past and the camera but never the cloud it is about to acquire.

\subsubsection{Causality of the camera cue.}
The scan mask is computed once per keyframe, before the returns of that keyframe are observed, and held over the intervening sweeps. The main evaluation uses the synchronized camera frame of each keyframe; we additionally evaluate a strict-causal variant that uses detections from the preceding keyframe, which are half a second stale and complete well before the current decision, and the vehicle study measures the real camera-to-decision lag.

\subsubsection{First-sighting events and metrics.}
A \emph{first sighting} is the first keyframe at which an annotated object within range $R$ appears in the scene, excluding objects present at scene start. The reported events are new entries; reappearances after annotation gaps are not scored separately, although they present the same absence of track support. For each policy we report: first-sighting recall, the fraction of such events detected at their first keyframe; time to first detection (TTFD) in keyframes, censored at $K$; and overall recall over all in-range object frames. Contrasts use paired per-scene bootstrap intervals, and results are stratified by range and azimuth.

\begin{remark}
\label{rem:protocol}
Throughout the experiments the grid is $A{=}512$ azimuth columns by $E{=}32$ elevation rows, matching the channel count of the rotating LiDAR in the evaluation data~\cite{caesar2020nuscenes}, with each azimuth column grouping a few neighboring firings.
The budgets are $\beta \in \{0.10, 0.20, 0.35\}$: the largest matches recent adaptive scanning, which cuts LiDAR energy by more than 65\%~\cite{adaptivelidar2026}, and the smaller two probe the tighter regime.
The diagnostic uses range $R{=}50$~m and a class-consistent 2~m center-distance match; the official nuScenes metrics apply the full protocol and are reported separately. TTFD is censored at $K{=}6$ keyframes, three seconds at 2~Hz~\cite{caesar2020nuscenes}.
The paired bootstrap uses $10^4$ per-scene resamples, a standard choice for percentile intervals~\cite{efron1994bootstrap}.
All values were fixed before the evaluation runs.
\end{remark}

\section{Method}
\label{sec:method}

Figure~\ref{fig:arch} shows the pipeline at one keyframe. The latest camera frame passes through a frozen 2D detector to give the detections $D_t$, and the track memory produces constant-velocity forecasts with their angular hulls $H_i$. The residual stage (Algorithm~\ref{alg:residual}) projects the forecasts into the image and keeps the detections that no forecast explains. The shared allocator (Algorithm~\ref{alg:allocator}) then fills the exploitation share from the hulls, the reserve from the residual set, and the remainder from a seeded floor to produce the mask $S_t$; the SBF stage (Algorithm~\ref{alg:sbf}) shrinks the hulls of low-criticality tracks before allocation. The mask is held over the sweeps, a frozen 3D detector runs on the retained points, and its detections update the track memory.

\subsection{Shared Budget Allocator}
\label{sec:allocator}

Every policy receives the same structure: an exploitation share of the budget, a reserve share, and a common random floor. Let $\rho \in (0,1)$ be the reserve fraction. The allocator first fills at most $(1-\rho)B$ cells from the exploitation sets, ordered nearest first and interleaved round-robin so that no single large object exhausts the share. It then fills the reserve layers in priority order, which keep at least $\rho B$ of capacity and inherit any unused exploitation; whatever remains is filled by a seeded random floor drawn with a seed shared by all policies at the same frame. This random floor is the only stochastic part of an otherwise deterministic pipeline. An unused reserve flows on to the floor, and a policy whose reserve is empty is exactly the history policy. Ties are broken by ascending cell index, and the total never exceeds $B$.

\begin{algorithm}[t]
\caption{Shared budget allocator (one keyframe, one policy)}
\label{alg:allocator}
\begin{algorithmic}[1]
\Require budget $B$; reserve fraction $\rho$; exploitation sets $\mathcal{H}$ (full hulls before shrunk hulls, nearest first within each); reserve layers $\mathcal{L}_1, \dots, \mathcal{L}_m$ (priority order); frame seed $s$
\State $S \gets \emptyset$
\State fill $S$ round-robin from $\mathcal{H}$ up to $\mathrm{round}((1-\rho)B)$ cells
\For{$k = 1, \dots, m$}
  \State fill $S$ round-robin from $\mathcal{L}_k$ up to $B - |S|$ cells
\EndFor
\State fill $S$ with the seeded random floor $\mathrm{Floor}(s)$ up to $B$ cells
\State \Return $S$
\end{algorithmic}
\end{algorithm}

The exploitation set of the base policy is built from track forecasts. Each tracked object $i$ with current position estimate is forecast one keyframe ahead by a constant-velocity rule, and its angular hull $H_i$ is the set of cells covered by the forecast box, dilated by a small azimuth margin and one elevation row to absorb forecast error. This base policy is our policy-level re-implementation of history-driven scanning~\cite{adaptivelidar2026}: dense attention on forecast regions, sparse floor elsewhere.

\subsection{CARE: the Camera-Residual Reserve}
\label{sec:care}

Let $D_t$ be the 2D boxes of the latest causally eligible camera frame from the frozen 2D detector. Each forecast is projected into the image; a detection $d \in D_t$ is \emph{explained} if the projected center of a same-class forecast falls inside its box. The residual set is
\begin{equation}
R_t = \{\, d \in D_t \;:\; \text{no same-class forecast center lies in } d \,\},
\label{eq:residual}
\end{equation}
and each residual box is mapped to a depth-free angular wedge $W(d)$ by back-projecting its corners and center to rays, with the same margins as the hulls; the wedge set $U = \{W(d) : d \in R_t\}$ is the residual reserve layer, and $M$ collects the wedges of the explained detections. CARE uses the wedge set $U$ as its camera reserve with strict priority over the shared floor; the explained set $M$ is kept for the priority-ordering variant and CARE+SBF, whose layers $(U, M)$ fill in that order.

\begin{algorithm}[t]
\caption{Residual reserve construction (CARE)}
\label{alg:residual}
\begin{algorithmic}[1]
\Require camera detections $D_t$; track forecasts $\mathcal{F}$; intrinsics and extrinsics
\State $U \gets \emptyset$; $M \gets \emptyset$
\For{$d \in D_t$}
  \If{some $f \in \mathcal{F}$ of the same class projects into the box of $d$}
    \State $M \gets M \cup \{W(d)\}$ \Comment{explained by memory}
  \Else
    \State $U \gets U \cup \{W(d)\}$ \Comment{residual: memory cannot explain it}
  \EndIf
\EndFor
\State \Return $U$ for CARE; $(U, M)$ for the priority-ordering variant and CARE+SBF
\end{algorithmic}
\end{algorithm}

Two design choices matter. First, matching uses the full forecasts, never the shrunk hulls of SBF, so forgetting can only reduce where the scanner looks and can never create a false residual. Second, the exploitation share is protected by its own cap and is filled first, so a burst of false 2D detections competes only for the reserve and never displaces tracked-object cells.

\begin{remark}
\label{rem:careparams}
The reserve fraction is $\rho{=}0.3$, fixed before the evaluation runs and swept in the ablation over $\{0.15, 0.3, 0.5\}$. The 2D detector is an off-the-shelf driving-domain YOLOX~\cite{ge2021yolox} at its public confidence threshold of $0.25$; the angular margin is two azimuth cells and one elevation row, absorbing forecast error of roughly one object radius at mid range; both are swept in the ablation. The 3D detector is a frozen CenterPoint~\cite{yin2021centerpoint} with its public configuration on the standard ten-sweep input. None of these values was tuned on the evaluation split.
\end{remark}

\subsection{Safety-Bounded Forgetting}
\label{sec:sbf}

The exploitation share itself contains waste: a parked car that has been re-scanned for many frames does not need its full hull every frame. SBF shrinks the hulls of tracks that the motion and guard rules classify as low criticality and leaves everything else untouched. A track is a candidate for forgetting only if it is receding faster than a threshold $\tau_{\mathrm{rec}}$ or quasi-static below a speed threshold $\tau_{\mathrm{st}}$, and, independently, only if its range exceeds a speed-dependent guard distance
\begin{equation}
d_{\mathrm{guard}} = k_{\mathrm{safe}} \left( s_0 + v_{\mathrm{ego}} \, T_h \right),
\label{eq:guard}
\end{equation}
so that nothing inside the ego interaction envelope is ever forgotten; every other track keeps its full forecast hull. Each consecutive safe frame increases a streak counter, and the hull margin decays with the streak on a fixed schedule that never drops the elevation pad. Because a real tracker carries localization noise, the shrunk hull keeps a residual margin of
\begin{equation}
m_{\mathrm{unc}} = \left\lceil \frac{\sigma_{\mathrm{pos}} / r}{\Delta\theta} \right\rceil
\label{eq:sigma}
\end{equation}
azimuth cells for a track at range $r$, where $\sigma_{\mathrm{pos}}$ is the position uncertainty and $\Delta\theta$ the cell width. The cells released by shrinking flow to the residual reserve through Algorithm~\ref{alg:allocator}, which couples retention and exploration in a single budget.

\begin{algorithm}[t]
\caption{Safety-bounded forgetting (SBF)}
\label{alg:sbf}
\begin{algorithmic}[1]
\Require tracks with two-frame history; ego speed $v_{\mathrm{ego}}$; streak counters
\For{each track $i$}
  \State $\mathrm{lowcrit}_i \gets (\mathrm{receding}_i \lor \mathrm{static}_i) \land \left(r_i > d_{\mathrm{guard}}\right)$ \Comment{Eq.~\eqref{eq:guard}}
  \State update streak: $c_i \gets c_i + 1$ if $\mathrm{lowcrit}_i$ else $0$
  \If{$\mathrm{lowcrit}_i$}
    \State shrink hull margin by the streak schedule, keep $m_{\mathrm{unc}}$ of Eq.~\eqref{eq:sigma}
  \Else
    \State keep the full forecast hull
  \EndIf
\EndFor
\State \Return exploitation sets (full hulls first, shrunk hulls second) for Algorithm~\ref{alg:allocator}
\end{algorithmic}
\end{algorithm}

\begin{remark}
\label{rem:sbfparams}
The guard follows published traffic rules and models: the headway term is $T_h{=}2$~s, the two-second rule, whose empirical support as a safety indicator is given by Vogel~\cite{vogel2003headway}; the standstill term is $s_0{=}2$~m, the jam distance of the intelligent driver model~\cite{treiber2000idm}; and the form of a speed-dependent safe envelope follows the responsibility-sensitive safety model~\cite{shalev2017rss}. The scale factor $k_{\mathrm{safe}}$ is swept in the ablation over $\{0.5, 1.0, 1.5, 2.0\}$ and fixed at $2.0$ beforehand; $\sigma_{\mathrm{pos}}$ is zero under ground-truth simulation and $0.4$~m on the vehicle. The motion thresholds are set against adult walking speed, 1.2 to 1.4~m/s~\cite{bohannon1997walking}: the static threshold $\tau_{\mathrm{st}}{=}0.5$~m/s stays below it, so a normally walking pedestrian is not classified as static, and the receding threshold $\tau_{\mathrm{rec}}{=}1$~m/s requires clear radial departure. At standstill the guard reduces to $k_{\mathrm{safe}} s_0$, which the vehicle experiment confirms exactly.
\end{remark}

\begin{remark}
\label{rem:positioning}
CARE is the recommended operating point and its first-sighting gains do not depend on SBF. The value of SBF lies in making the retention side of the budget explicit and safe: the guard evidence in Section~\ref{sec:ablations} shows that forgetting without the guard significantly harms near-field recall, whereas the guarded default releases budget with no significant loss at the tightest budget.
\end{remark}

\subsubsection{Failure handling.}
When the camera produces no eligible detections, $R_t$ is empty and the reserve returns to the shared floor, so CARE equals the history policy by construction; Section~\ref{sec:vehicle} documents this fallback under a real camera failure. False 2D detections cannot displace the exploitation cells, which are filled first under their own cap.

\section{Experiments}
\label{sec:exp}

\subsection{Setup}
\label{sec:setup}

\subsubsection{Data and detectors.}
The primary evaluation uses the nuScenes validation split~\cite{caesar2020nuscenes} with 150 scenes, 6{,}019 keyframes, and 4{,}148 first-sighting events; 20 disjoint scenes only train the learned-allocation baseline. Detections of the frozen YOLOX are precomputed on all six cameras, and all emulation, protocol, and parameter values follow Remarks~\ref{rem:protocol}--\ref{rem:sbfparams}. Both detectors and the ground truth share the five driving classes of the LiDAR head (car, truck, bus, bicycle, pedestrian); the other 2D classes are dropped, and the remaining five nuScenes classes lie outside the evaluation.

\subsubsection{Compared policies.}
All policies run through Algorithm~\ref{alg:allocator} at identical budget, identical floor seeds, and identical detector. A floor seed fixes the random fill of the leftover budget, so all policies draw their floor with the same seed at a frame. We average history, the all-camera reserve, and CARE with and without SBF over three floor seeds, with per-seed values within 0.4 points of the mean, far below the differences we report; the remaining policies, the official metrics, and the ablations use a single seed. The cell-budget policies are: \emph{history}, our policy-level re-implementation of adaptive scanning~\cite{adaptivelidar2026}; \emph{history+uniform}, which spends the reserve on a fixed uniform pattern; \emph{all-camera}, which reserves for every camera detection and thus removes only the residual filter; \emph{uncertainty}, a light-curtain style reserve on cells that were occupied but undetected in the previous frame~\cite{ancha2020lightcurtains}; \emph{beam-reduce}, which keeps the elevation rows nearest the horizon~\cite{wei2022lidardistill}; a \emph{learned} scorer trained on the calibration scenes~\cite{lang2020samplenet}; and \emph{CARE} with and without SBF. As non-causal downsampling references at matched point fraction we report random point sampling and, at small scale, farthest point sampling~\cite{qi2017pointnetpp}; these see the full cloud and save no sensing.

\subsection{Main Results}
\label{sec:mainresults}

\begin{figure}[t]
\centering
\includegraphics[width=0.98\textwidth]{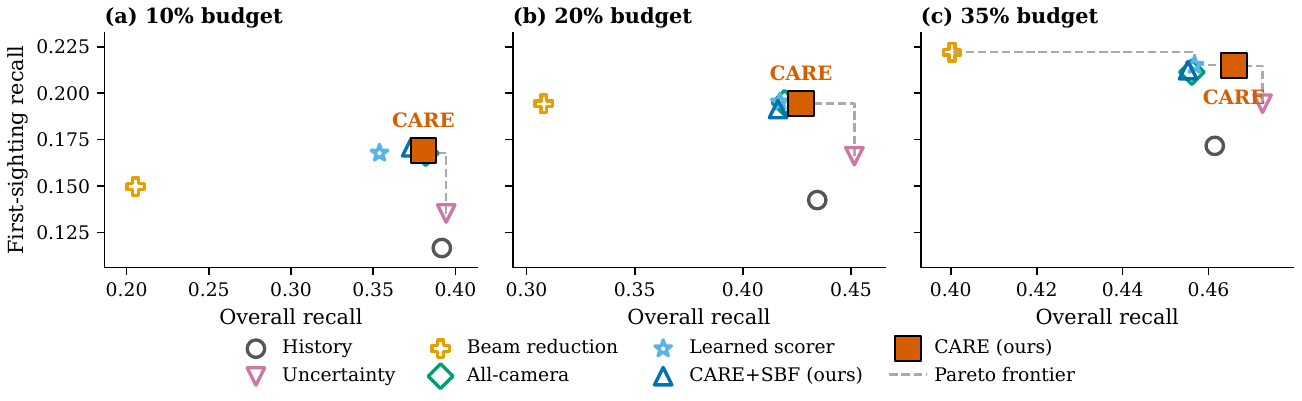}
\caption{First-sighting recall versus overall recall at the (a) 10\%, (b) 20\%, and (c) 35\% budgets on nuScenes validation. The gray staircase traces the Pareto frontier; CARE sits at its knee in every panel, and Table~\ref{tab:main} gives the paired contrasts.}
\label{fig:main}
\end{figure}

\begin{figure}[t]
\centering
\includegraphics[width=0.98\textwidth]{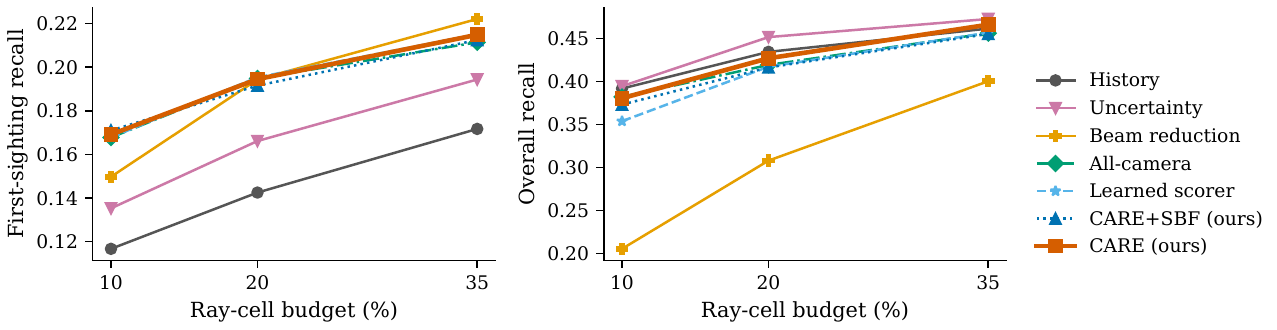}
\caption{First-sighting recall (left) and overall recall (right) versus ray-cell budget for CARE, CARE+SBF, and the policy-level re-implementations: history-driven scanning~\cite{adaptivelidar2026}, the uncertainty reserve~\cite{ancha2020lightcurtains}, beam reduction~\cite{wei2022lidardistill}, the all-camera reserve, and the learned scorer~\cite{lang2020samplenet}.}
\label{fig:levels}
\end{figure}

CARE raises first-sighting recall over the history policy by 5.2, 5.2, and 4.3 points at the 10\%, 20\%, and 35\% budgets (Table~\ref{tab:main}, Figure~\ref{fig:main}), all intervals excluding zero; equal scene weights give 5.3, 5.2, and 4.2 points, so event-rich scenes do not drive it.
\begin{table}[t]
\centering
\footnotesize
\caption{Paired per-scene bootstrap contrasts, CARE minus the named policy, in points of recall with 95\% confidence intervals; positive favors CARE; bold marks intervals excluding zero in favor of CARE. First-sighting contrasts not shown are statistically tied, except beam reduction at 10\%, where CARE leads by 2.0 points; the significant overall-recall contrasts against history and the uncertainty reserve appear in the text. Cells use one decimal, or two where an endpoint would round to zero; the identical history cells at 10\% and 20\% differ before rounding. Absolute levels appear in Figure~\ref{fig:levels}.}
\label{tab:main}
\setlength{\tabcolsep}{3pt}
\begin{tabular}{lccc}
\toprule
Contrast & 10\% & 20\% & 35\% \\
\midrule
\multicolumn{4}{l}{\emph{First-sighting recall (points)}} \\
vs. History~\cite{adaptivelidar2026} & \textbf{+5.2 [4.3,6.2]} & \textbf{+5.2 [4.3,6.2]} & \textbf{+4.3 [3.5,5.1]} \\
vs. Uncertainty reserve~\cite{ancha2020lightcurtains} & \textbf{+3.5 [2.6,4.4]} & \textbf{+2.7 [1.7,3.8]} & \textbf{+2.1 [1.1,3.1]} \\
\addlinespace
\multicolumn{4}{l}{\emph{Overall recall (points)}} \\
vs. Beam reduction~\cite{wei2022lidardistill} & \textbf{+17.4 [15.9,19.0]} & \textbf{+11.5 [10.4,12.7]} & \textbf{+6.4 [5.4,7.5]} \\
vs. Learned scorer~\cite{lang2020samplenet} & \textbf{+2.4 [1.9,3.0]} & \textbf{+0.7 [0.4,1.0]} & \textbf{+0.6 [0.3,0.9]} \\
vs. All-camera reserve & $-$0.13 [$-$0.26,$-$0.01] & \textbf{+0.5 [0.3,0.8]} & \textbf{+0.7 [0.4,1.0]} \\
\bottomrule
\end{tabular}
\end{table}

\begin{table}[tb]
\centering
\footnotesize
\caption{Official nuScenes metrics under the ray-cell budgets. Because these metrics average over every in-range object frame, and most of those frames belong to objects the tracker already follows, a gain that is confined to first sightings carries little weight in the average, and CARE stays close to the history policy on mAP while trailing it on NDS. This dilution is the reason we measure first sightings separately. The detector head emits five of the ten protocol classes, so the absolute values are not comparable with standard ten-class reports.}
\label{tab:official}
\setlength{\tabcolsep}{4pt}
\begin{tabular}{lcccccc}
\toprule
& \multicolumn{3}{c}{mAP} & \multicolumn{3}{c}{NDS} \\
\cmidrule(lr){2-4}\cmidrule(lr){5-7}
Policy & 10\% & 20\% & 35\% & 10\% & 20\% & 35\% \\
\midrule
History & 0.076 & 0.097 & 0.110 & 0.131 & 0.164 & 0.169 \\
All-camera & 0.070 & 0.090 & 0.107 & 0.124 & 0.130 & 0.162 \\
CARE+SBF (ours) & 0.066 & 0.088 & 0.106 & 0.120 & 0.128 & 0.162 \\
\textbf{CARE (ours)} & 0.068 & 0.091 & 0.111 & 0.124 & 0.149 & 0.164 \\
\bottomrule
\end{tabular}
\end{table}

A full scan reaches 0.245 first-sighting and 0.494 overall recall; CARE at 35\% recovers 88\% and 94\% of these ceilings against 70\% and 93\% for history, so it closes most of the allocation-controlled gap; the low ceiling reflects a detector and return limit.
The all-camera control shows that the camera cue, not the history state, drives the first-sighting gain: the two are tied on first sightings at all three budgets. Residual filtering keeps the reserve on objects the memory cannot explain instead of on tracked ones; on average the two reserves differ on overall recall by only $-0.1$, $+0.5$, and $+0.7$ points, a small deficit at 10\% and small gains at the larger budgets, each interval excluding zero (Table~\ref{tab:main}). The near-tie is mechanical: the filter removes about a thousand cells of explained wedges per keyframe, but at 10\% the unexplained wedges alone fill the reserve on 79\% of keyframes, and the realized allocation changes on only 17\% of keyframes, against 68\% at 35\%. The value appears where scenes are crowded: in the densest third of the scenes, ranked by tracked-object count, CARE gains 0.7 and 1.0 points of overall recall over the all-camera reserve at 20\% and 35\%. The dense-minus-sparse difference of 0.7 and 1.3 points excludes zero, and the same pattern appears when scenes are ranked by explained detections instead. Sparse scenes are tied, and the stratified first-sighting contrasts are tied apart from a 1.3-point loss in the middle third at 20\%; Section~\ref{sec:carla} adds the closed-loop counterpart at scarce budgets. The depth-free class gate falsely explains 32.5\% of new entries with a 2D detection, yet reserving those directions recovers only four events at 10\%: the detector misses 85\% of them even when their directions are scanned, and CARE already detects nearly all of the remainder through the floor and neighboring coverage, so the ambiguity is nearly outcome-neutral.
Against the history policy, the first-sighting gain is not free at tight budgets: in the paired per-scene contrast used for all intervals, overall recall moves by $-0.8$, $-0.5$, and $+0.6$ points at 10\%, 20\%, and 35\%, a budget-dependent Pareto choice that we report as such rather than as a uniform win.
The learned scorer matches CARE on first-sighting recall but loses 0.6 to 2.4 points of overall recall at every budget (Table~\ref{tab:main}); that a learned allocator does not beat the rules also indicates the history baseline, our concept-level re-implementation of a learned region predictor, is not understating such methods here.
An oracle variant sharpens this point: replacing the forecast of every tracked object with its true next position, an upper bound on any learned region predictor, moves first-sighting recall by at most 0.5 points in a single-seed diagnostic run, and every interval contains zero. Within the same run CARE stays 3.7 to 4.7 points above this oracle, so what it adds is information about objects that have no track yet, not a better forecast. The oracle raises overall recall by 0.5 to 0.7 points, as expected from perfect placement on tracked objects. Descriptively, on a single seed CARE lowers the mean censored TTFD of the history policy from 3.76 to 3.64, 3.60 to 3.47, and 3.45 to 3.33 keyframes, a shift of about 0.12 keyframes (60~ms); no interval claim is made for TTFD. The effect-size gate fixed before the runs, five points of first-sighting recall, is met at 10\% and 20\%; the 35\% gain stays positive but below it.
Under the official protocol (Table~\ref{tab:official}), mAP shows the same transition, with CARE below history at 10\% and 20\% and slightly above at 35\%, while NDS stays below history throughout, since these frame-averaged metrics dilute a first-sighting gain and thereby motivate the separate protocol.

\subsection{Baseline Comparison}
\label{sec:baselines}

The uncertainty reserve confirms the central distinction. It attains the best overall recall of all policies, exceeding CARE by 1.0, 2.2, and 0.5 points with intervals excluding zero, because it re-examines regions that returned points without a confident detection. Yet it trails CARE on first sightings by 3.5, 2.7, and 2.1 points at all three budgets (Table~\ref{tab:main}), since an object that has never returned a point creates no uncertainty to react to.
Beam reduction shows the opposite behavior: its first-sighting recall is statistically tied with CARE at 20\% and 35\% and 2.0 points behind at 10\%, while it gives up 17.4, 11.5, and 6.4 points of overall recall (Table~\ref{tab:main}), so CARE gives the better joint trade-off.
The history+uniform control falls below plain history at every budget, so an undirected reserve costs more than it returns.
Across the comparisons (Figure~\ref{fig:main}), CARE combines the best or statistically tied first-sighting recall at every budget with the highest overall recall among the policies that match its first sightings at 20\% and 35\%, conceding at most 2.2 points of overall recall to the uncertainty reserve; this joint behavior is the design goal of the residual reserve.

\subsection{Ablations}
\label{sec:ablations}

Each mechanism is swept on a fixed quarter of the validation scenes, every fourth scene, 38 scenes with 925 events; differences below about 1.5 points are within the subset noise and read as ties.
The reserve fraction needs no tuning: first-sighting recall moves by at most one point across the swept fractions, since the exploitation cap protects tracked objects and unused reserve returns to the floor.
Priority ordering is structural: on the full split the variant changes first-sighting recall by at most 0.2 points, with overall recall of 0.380, 0.418, and 0.457.
The forgetting knobs are insensitive on first sightings: the guard scale, the localization margin, and the four decay schedules stay within the subset noise, because at urban speeds the guard of Eq.~\eqref{eq:guard} already covers most nearby objects; the pre-registered defaults of Remark~\ref{rem:sbfparams} are kept.
The guard itself, however, is load bearing. At the 10\% budget, relative to CARE without forgetting, unguarded forgetting loses 1.8 points of near-field recall on objects within 20~m, interval $[-2.9, -0.5]$, a quarter-scale guard loses 1.5 points $[-2.7, -0.2]$, and the default shows no significant loss, $-0.5$ $[-1.7, +0.4]$, while releasing 24 cells per keyframe. At 20\% even the default pays 3.1 points $[-4.7, -1.5]$, so forgetting is a tool for the tightest budgets.

Three further sweeps probe the camera pathway.
Delaying the camera detections by one full keyframe, so the reserve only sees half-second-old evidence, still recovers 4.6, 5.3, and 2.6 points over history: same-frame camera input is not required.
Halving or doubling the angular margin keeps the paired gap between 3.3 and 6.3 points, and raising the 2D confidence threshold from 0.25 to 0.5 costs one point at the tightest budget and nothing above; all three sweeps leave the policy ordering unchanged and are descriptive.

\subsection{Vehicle Case Study}
\label{sec:vehicle}

\begin{figure}[tb]
\centering
\includegraphics[width=0.72\textwidth]{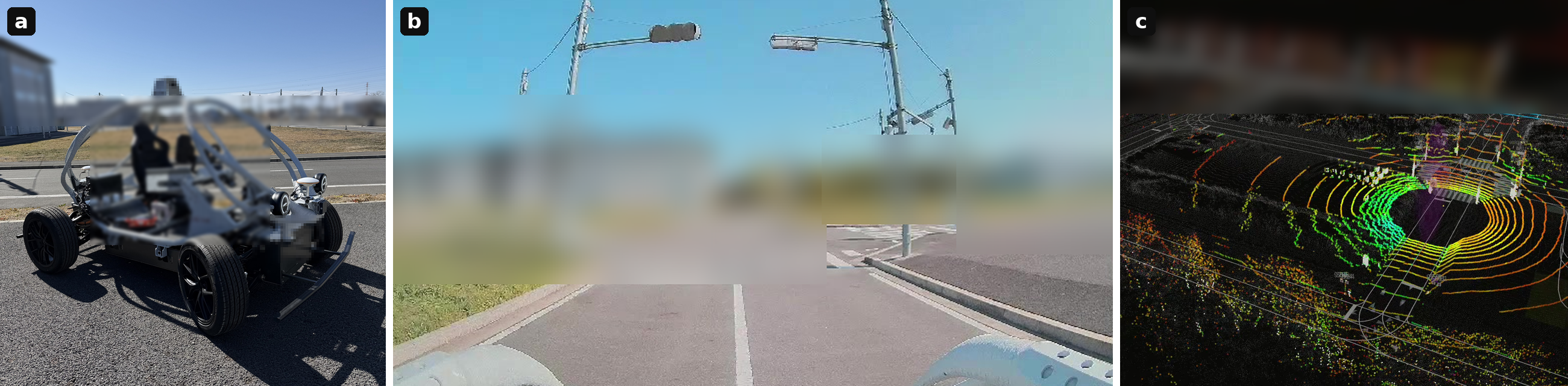}
\caption{The vehicle platform: (a) drive-by-wire chassis with LiDAR, forward camera, and GNSS; (b) camera view and (c) LiDAR cloud from a separate drive.}
\label{fig:vehicle}
\end{figure}

The platform is a drive-by-wire vehicle with one rotating LiDAR and one forward camera (Figure~\ref{fig:vehicle}); we analyze a driving sequence with the camera operating and a garage session during which the camera failed and delivered no frames.

On the driving sequence the pipeline runs end to end with real YOLOX detections and ego-motion compensated forecasts. The reserve uses the newest camera frame at or before each decision, rotating wedge directions by the ego yaw change over the gap; this evidence is 0.11 to 0.14~s old on average against the 0.30~s keyframe spacing, and 0.42~s at the 95th percentile, a staleness the delayed-camera sweep covers. Across 104 new-entry events CARE detected the entry earlier than the history policy on 17 and later on 8, with the rest tied. Matched detections were rare, 19 keyframes of 648: at first appearance the camera mostly sees objects the tracker does not hold, so CARE and the all-camera reserve are nearly identical here; intrinsics are approximate.

In the garage session the perception path had a steady-state compute p99 of 97.9~ms against a 100~ms budget over 1{,}000 frames; the 2D stage used synthetic inputs because the camera recorded no images; acquisition and transport lie outside the boundary. At standstill the guard of Eq.~\eqref{eq:guard} equaled $k_{\mathrm{safe}} s_0 = 4.00$~m on all 570 keyframes, and the margin of Eq.~\eqref{eq:sigma} kept the net release near zero, the intended conservative behavior.

With the camera absent, CARE, all-camera, and history produced identical allocations on all 37 stable new-entry events, while newly appearing clusters received as few as 1 to 5 cells at appearance against 117 to 123 inside established coverage: the diagnosed under-allocation is visible on the vehicle. Several close or side entries were missed by every policy, so equality under the fallback does not imply detection safety; all vehicle results are descriptive.

\begin{figure}[tb]
\centering
\begin{minipage}[c]{0.40\textwidth}\centering
\includegraphics[width=\textwidth]{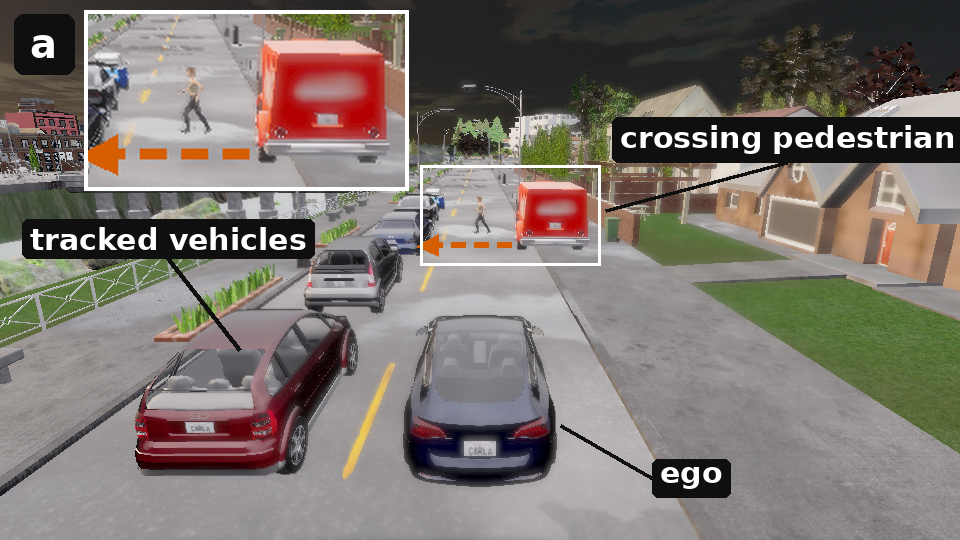}
\end{minipage}\hfill
\begin{minipage}[c]{0.57\textwidth}\centering
\includegraphics[width=\textwidth]{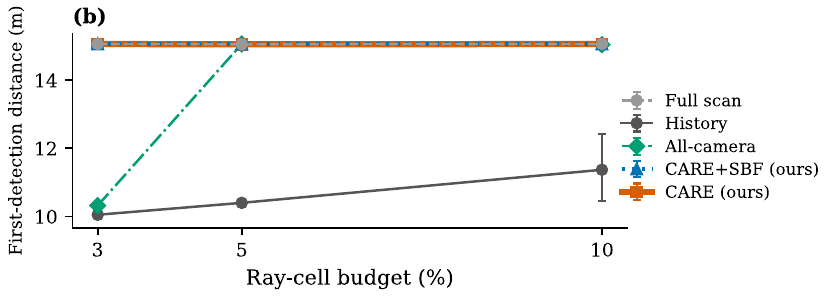}
\end{minipage}
\caption{The dense CARLA scene. (a) The ego follows a street lined with tracked vehicles; the pedestrian crossing from behind the parked truck (dashed track, magnified inset) is the first sighting the reserve must catch. (b) Median first-detection distance of the crosser, five seeds per point, interquartile bars. The all-camera reserve collapses at 3\%, spread over the five tracked vehicles; CARE and CARE+SBF match the full-scan bound throughout, so the three top curves coincide.}
\label{fig:carla}
\end{figure}

\subsection{Closed-Loop Simulation Study}
\label{sec:carla}

To test closed-loop consequences we add a controlled CARLA study~\cite{dosovitskiy2017carla}: a fixed controller brakes on detected objects, the same allocator masks the cloud, the camera cue uses ground-truth boxes with injectable false negatives and latency, and detection needs enough masked returns. Scenario physics were fixed before any policy ran; all 245 episodes over five seeds are identical across policies.

When a pedestrian darts out from behind a parked vehicle, at the 10\% budget the camera policies detect it at about 15~m, brake in every episode, and pass at 5.7~m or more; history detects at 11.9~m, brakes in three of its five episodes, and passes at 4.0~m. In the dense scene of Figure~\ref{fig:carla}, at 3\% CARE detects the crosser at 15.1~m, the full-scan bound, while the all-camera reserve spreads over the tracked vehicles and detects at 10.3~m; they converge from 5\% upward. A vehicle cut-in shows no separation, since a large vehicle returns enough points even when starved: the camera cue matters for small or sparse first sightings. At 10\%, false-negative rates up to 0.4 and a two-frame latency leave the ordering unchanged. No episode collides under the fixed physics, so we report detection and braking; the proxy upper-bounds a learned detector, and we make no deployment claim.

\section{Conclusion}
\label{sec:conclusion}

History-driven adaptive LiDAR scanning under-samples exactly the objects it has never seen. This paper measured that failure with a leakage-free protocol and repaired it with CARE, a training-free reserve for camera detections that memory cannot explain. On nuScenes validation CARE improves first-sighting recall over history at every budget with intervals excluding zero; the camera cue drives this gain; residual filtering costs about 0.1 points of overall recall at the tightest budget and pays off in crowded scenes. The mechanism runs on a real vehicle and, in simulation, sees an occluded pedestrian earlier and brakes more reliably; the evidence is limited to one frozen detector pair, ray-cell emulation, and new-entry events; steerable hardware is the next step.

\bibliographystyle{splncs04}
\bibliography{main}

\end{document}